\documentclass[journal, 10pt,twoside,web]{ieeecolor}

\usepackage{amsmath}

\usepackage{amssymb}
\usepackage{float}

\usepackage{dblfloatfix} 
\usepackage{multirow}
\usepackage{generic}
\usepackage{amsmath,amssymb,amsfonts}
\usepackage{algorithmic}
\usepackage{graphicx}
\usepackage{textcomp}

\usepackage{graphicx}
\usepackage{subfigure}
\usepackage{epsfig} 

\usepackage{lcsys}

\usepackage[style=ieee]{biblatex}
\renewbibmacro*{in:}{}
\AtEveryBibitem{%
  \clearfield{url}%
  \clearfield{doi}%
  \clearfield{isbn}%
  \clearfield{issn}%
  \clearfield{month}%
  \clearfield{note}%
  \clearfield{volume}%
  \clearfield{number}%
  \clearfield{pages}%
  \clearlist{language}%
  \clearfield{location}%
  \clearlist{location}%
  \clearfield{editor}
  \clearlist{editor}
  \clearname{editor}
  \clearfield{publisher}
  \clearlist{publisher}

}
\begin{document}

\def\BibTeX{{\rm B\kern-.05em{\sc i\kern-.025em b}\kern-.08em
    T\kern-.1667em\lower.7ex\hbox{E}\kern-.125emX}}
\markboth{\journalname, VOL. XX, NO. XX, XXXX 2017}
{Author \MakeLowercase{\textit{et al.}}: Preparation of Papers for IEEE Control Systems Letters (August 2022)}

\title{Gradient based Bilevel for Inverse Optimal Control,
a Riemannian approach}

\author{
Ahmed-Manaf Dahmani$^{1,2}$,
Vincent Bonnet$^{3}$,
David Daney$^{1}$,
François Charpillet$^{2}$
\thanks{$^{1}$ INRIA Bordeaux, Université de Bordeaux, Talence, France.}
\thanks{$^{2}$ INRIA Nancy, Université de Lorraine, Villers-lès-Nancy, France.}
\thanks{$^{3}$ LAAS-CNRS, Université de Toulouse, CNRS, Toulouse, France.}
}

\maketitle
\thispagestyle{empty}

\begin{abstract}
Inverse Optimal Control (IOC) aims to recover the cost function that explains observed trajectories as solutions of an optimal control problem. Classical IOC formulations rely on bilevel optimization, which repeatedly solves a nested optimal control problem and quickly becomes computationally prohibitive for realistic systems. Recent projection-based approaches offer a promising alternative but suffer from numerical instability when solved with gradient-based methods due to violations of standard constraint qualifications.

In this paper, we show that these difficulties stem from the geometric structure of the IOC feasible set. We demonstrate that the set of trajectories satisfying the optimality conditions naturally forms a manifold and reformulate IOC as an optimization problem on this manifold. Based on this insight, we propose a Riemannian Inverse Optimal Control (RIOC) method that projects observed trajectories onto the manifold of optimal solutions while preserving feasibility by construction. Experiments on real human arm trajectories show that the proposed method achieves comparable or better reconstruction accuracy than classical bilevel IOC while reducing computation time by about a factor of four. These results highlight the potential of geometric optimization methods to improve the scalability and reliability of IOC for robotics and human motion analysis.
\end{abstract}

\begin{IEEEkeywords}
 Inverse Optimal Control, Bilevel, Riemannian Optimization
\end{IEEEkeywords}

\section{Introduction}

Inverse Optimal Control (IOC) aims at recovering the cost function that explains observed trajectories as solutions of an optimal control problem. In robotics and human–robot interaction \cite{berret_evidence_2011, becanovic_force_2023, panchea_human_2018, mainprice_predicting_2015, englert_inverse_2018, lin2021} to replicating it from observations, IOC and more recently Inverse Reinforcement Learning (IRL) \cite{finn2016} are widely used to model human motion from demonstrations, with applications ranging from motion prediction to skill transfer. In these settings, observed motions are assumed to reflect consistent trade-offs between task performance and physical or kinematic effort, which can be compactly represented through interpretable cost functions.

Classical IOC approaches rely on a bilevel formulation \cite{panchea_inverse_nodate}, where cost parameters are identified by repeatedly solving a nested optimal control problem while minimizing the discrepancy with observed trajectories. Although principled, bilevel IOC is computationally expensive and difficult to scale to high-dimensional systems. IRL offers a probabilistic alternative that alleviates some nested optimization, but its performance strongly depends on the quality of the trajectory-space approximation and often requires extensive sampling around demonstrations \cite{finn2016}. As a result, both IOC and IRL face significant challenges when applied to real-world robotic or human motion data \cite{lin2021}. To address computational limitations, several approximate IOC methods have been proposed, including formulations based on Karush–Kuhn–Tucker (KKT) residuals or local linearizations. While these approaches yield solutions consistent with first-order optimality conditions, they do not explicitly enforce proximity to the observed trajectories, which is often the primary objective in practical applications.

Recently, a geometric interpretation of IOC and IRL solvers has been introduced, viewing cost identification as a projection problem in trajectory space. This unified perspective led to Projected IOC (PIOC) \cite{colombel_holistic_2023,colombel_reliability_2022}, which formulates cost inference as an orthogonal projection onto singularity curves associated with optimality conditions. Although theoretically appealing, PIOC has shown poor numerical reliability when solved using standard gradient-based optimization methods.

In this paper, we show that the instability of PIOC originates from systematic constraint violations that make the problem ill-suited for Euclidean optimization. We analyze the underlying geometric structure of the feasible set and show that it naturally admits a Riemannian manifold formulation. Based on this insight, we propose a new IOC resolution method using Riemannian optimization that enforces feasibility by construction and ensures reliable convergence with first-order solvers.

The proposed method, referred to as Riemannian IOC (RIOC), preserves the geometric benefits of projection-based IOC while avoiding the numerical issues of PIOC. We demonstrate on real human motion data that RIOC converges faster than bilevel IOC while maintaining accuracy, providing a scalable and reliable alternative for both IOC and IRL-based learning from demonstrations.

In this context, the main contributions of this work are :
\begin{itemize}
    \item Explaining the unsuitability of gradient based solvers in solving of the PIOC formulation of the IOC; 
    \item Making use of the holistic view of the IOC provided by the PIOC formulation to propose a new Riemannian optimization based IOC method;
    \item Comparing the performance of the new Riemannian method the Bilevel using real human data.
\end{itemize}
\section{Background}\label{sec:background}
\subsection{Direct Optimal Control} \label{sec:doc}
Direct Optimal Control (DOC) is the application of optimization problems to system control. It is widely used in real-world applications such as decision making, path prediction and planning \cite{englert_inverse_2018,mainprice_predicting_2015, puydupin-jamin_convex_2012}. The DOC problem consists in minimizing a parameterized cost function $C(s,\omega)$ where $s$ is the trajectory and optimization variable and $\omega$ is the cost function parameter. It can be subject to equality constraints $g(s)=0$ and inequality constraints $h(s) \ge 0$. The parameterized cost function is often a weighted  linear combination of a base of cost functions $C(s,\omega) = \sum_{k=1}^{n_c}\omega_k C_k(s)$ with $n_c$ the number of considered cost functions. In the general case the DOC is formulated as: 
\begin{equation}
        s^{*}=\mathop{\mathcal{D}}(\omega) = \operatornamewithlimits{argmin}_{s} C(s,\omega)\hspace{0.25cm}\text{s.\,t.\,} g(s)=0, h(s) \ge 0
        \label{eq:doc_lin}
\end{equation}
DOC are often solved using primal-dual methods, such as Interior Point Method (IPM) \cite{curtis_interior-point_2010}, for their robustness. IPM tries finding an optimal solution $s^{\star}$ by solving the a system of equations known as the Karush–Kuhn–Tucker (KKT) conditions. Assuming $n_g$ the number of equality constraints and $n_h$ the number of inequality constraints, the KKT conditions are as shown in \eqref{eq:kkt}. $\lambda$ and $\mu$ are called the KKT multipliers.
\begin{small}
\begin{equation}
\begin{split}
\exists \lambda^\star \in \mathbb{R}^{n_g}, \exists \mu^\star \in \mathbb{R}^{n_h},\hspace{5.2cm}&\hspace{-0.4cm}\\
\begin{cases}
    \frac{\partial C}{\partial s}^T(s^{\star}) \omega + 
    \frac{\partial g}{\partial s}^T(s^{\star}) \lambda^\star +
    \frac{\partial h}{\partial s}^T(s^{\star}) \mu^\star 
  = 0 \hspace{0.1cm}  \text{(Stationarity)}&\\
    g(s^{*}) = 0,\quad h(s^{*}) \ge 0 \hspace{2.15cm}  \text{(Primal feasibility)}&\\
    \mu^\star \ge 0 \hspace{4.6cm}  \text{(Dual feasibility)}&\\
    h(s^{*})^T\mu^\star = 0   \hspace{2.25cm} \text{(Complementary slackness)}&
    \end{cases}
\end{split}
    \label{eq:kkt}
\end{equation}
\end{small}

IPM solvers handle inequality constraint by adding a barrier function to the cost function $F(s) = C(s)^T\omega -\mu_{\text{target}}\sum_{i=1}^{n_h}ln(h(s))$. The parameter $\mu_{\text{target}}$ controls the precision of the approximation of the solution. The smaller it is the closer the result is to the real solution.  The KKT conditions imply that the following error vector $E(s^\star, \omega, \lambda^\star, \mu^\star)$ must be zero.
{\footnotesize
\begin{equation}
E(s^\star, \omega, \lambda^\star, \mu^\star) =\begin{pmatrix}
    \frac{\partial C}{\partial s}^T(s^{\star}) \omega + 
    \frac{\partial g}{\partial s}^T(s^{\star}) \lambda^\star +
    \frac{\partial h}{\partial s}^T(s^{\star}) \mu^\star\\
    g(s^{*})\\
    h(s^\star)\circ \mu^\star - \mu_{\text{target}} * 1_{n_h}
\end{pmatrix}
    \label{eq:kkt2}
\end{equation}
}

Differentiating this system of equations results in a linear system that the change in each variable must verify. Introducing the Lagrangian $\mathcal{L} = \sum_{i=1}^{n_c} \omega_i    C +  \sum_{i=1}^{n_g} \lambda_i    g +  \sum_{i=1}^{n_h} \mu_i    h$ we get:
{\footnotesize
\begin{equation}
    \begin{pmatrix}
        \frac{\partial^2 \mathcal{L}}{\partial s^2} &  \frac{\partial g}{\partial s}^T & \frac{\partial h}{\partial s}^T\\
        \frac{\partial g}{\partial s} & 0 & 0\\
       diag(\mu)\frac{\partial h}{\partial s}  & 0 & h
    \end{pmatrix}\begin{pmatrix}
        \Delta s\\ \Delta \lambda \\ \Delta \mu
    \end{pmatrix}
    = -E
    \label{eq:kkt3}
\end{equation}
}

The DOC problem is assumed to respect conditions known as constraint qualifications \cite{nocedal_theory_2006} for \eqref{eq:kkt3} to be solvable. IPM solvers assume the Linear Independence Constraint Qualification (LICQ) stating that the gradients of the equality constraints and active inequality constraints are linearly independent near the optimal solution. This is the strongest and most commonly assumed constraint qualifications for non-linear optimization.

In practice, IPM solvers like the popular IPOPT\cite{wachter_interior_2002} introduce slack variables and solve an equivalent symmetric linear system to reduce the dimension of search space and improve numerical stability. 
\subsection{Inverse Optimal Control}
The IOC aims to find the value of the weight vector $\omega$ that best matches the observed trajectory. A commonly used approach is minimizing a distance between the observed trajectory $s_{\text{obs}}$ and the result of the DOC as shown in \eqref{eq:bilevel}. A normalization constraint $\|\omega\|_p = 1$ is added to ensure uniqueness of $\omega$. This method consists of a double optimization loop and is commonly referred to as the Bilevel approach\cite{albrecht_bilevel_2012, panchea_human_2018}.
{\footnotesize
\begin{equation}
\begin{split}
        \omega^{*}=\mathop{\mathcal{I}}(s_{obs}) =& \operatornamewithlimits{argmin}_{\omega} \| s_{obs} - \mathop{\mathcal{D}}(\omega)\|^2\\
        &\quad\text{s.\,t.}\,\quad \omega > 0,\quad \|\omega\|_p = 1
    \end{split}
    \label{eq:bilevel}
\end{equation}
}
Since there is no clear structure on the solution space of the DOC, this optimization problem is often solved using derivative free methods such as Powell's method\cite{powell_efficient_1964} or COBYLA\cite{powell_direct_1994}. These methods require numerous DOC calls to approximate the step direction, resulting in a computationally slow method. This pushed for a search for faster IOC resolution methods.
\subsection{Holistic view of IOC}
\begin{figure}[H]
    \centering

        \includegraphics[width=0.32\textwidth]{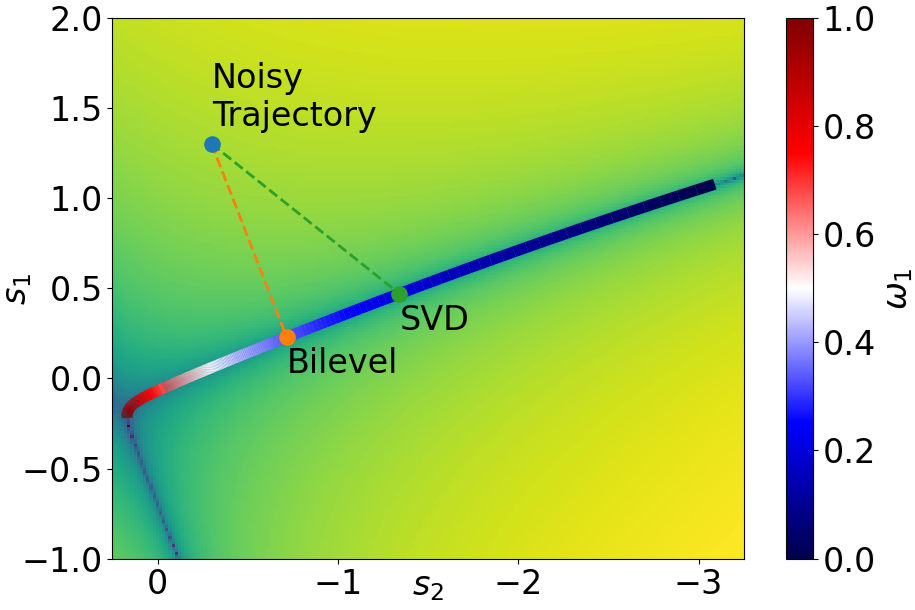}
            
    \caption{ $\omega_1$ as parameterization of SC and IOC methods as projections}
    
    \label{fig:omega_dist} 
\end{figure}
In the search for faster IOC resolution methods, many approximate methods were explored \cite{panchea_human_2018, englert_inverse_2018,colombel_reliability_2022}. These methods minimize the KKT conditions. However, minimizing a distance to the observed trajectory presents a useful property in practical settings. The search for a fast resolution method conserving this property has led Colombel et al. \cite{colombel_holistic_2023} to propose a so-called holistic view of the IOC by exploring the structure of the solution space of the DOC. The paper defines the Singularity Curve (SC) as the set of trajectories verifying the KKT stationarity condition \eqref{eq:kkt}. This set is often a smooth continuous shape as shown in Fig. \ref{fig:omega_dist}, allowing for the description of DOC as a smooth map from $\mathbb{R}^{n_c}$ to $SC$.  Inversely, different resolution methods of the IOC represent different projections on the SC. When a trajectory is already optimal, it lies on the SC and all projections coincide. However, in real-world applications, trajectories are never truly optimal due to system noise and measurement error. In this case the result of the IOC methods would differ. With this perspective in mind, a projection that minimizing a distance to the observation is an orthogonal projection on the SC. Fig. \ref{fig:omega_dist} shows an illustration of an orthogonal and SVD based\cite{colombel_reliability_2022}. In an attempt to formulate the IOC problem as a one level optimization problem performing an orthogonal projection on the SC, the stationarity condition \eqref{eq:kkt} is written in matrix form as in \eqref{eq:kkt_lin}. The existence of a vector verifying the stationarity equation implies the singularity of the matrix $J(s^{\star})$, resulting in the description of $SC = \{s^{\star} \in \mathbb{R}^{m} | \mathop{det}(J(s^{\star})^TJ(s^{\star})) = 0\}$.
{\footnotesize
\begin{equation}
        \underbrace{
    \begin{pmatrix}
    \frac{\partial C}{\partial s}^T(s^{\star}) &
    \frac{\partial g}{\partial s}^T(s^{\star}) &
    \frac{\partial h}{\partial s}^T(s^{\star})
    \end{pmatrix}
    }_{J(s^{\star}) = \begin{pmatrix}
    J_\omega (s^{\star})&
    J_\lambda (s^{\star})&
    J_\mu(s^{\star})
    \end{pmatrix}}
    \begin{pmatrix}
    \omega\\
    \lambda\\
    \mu
    \end{pmatrix} = 0 
    \label{eq:kkt_lin}
\end{equation}
}
A recent method called the Projected IOC (PIOC) \cite{colombel_reliability_2022} leverage this alternate description of the $SC$ is then formulated as shown in \eqref{eq:PIOC}. This formulation of the IOC as a one level optimization problem with differentiable constraints should allows in principle for a faster resolution time.
\begin{small}
\begin{equation}
    \mathop{\mathcal{I}_{\text{PIOC}}}(s_{obs}) = \operatornamewithlimits{argmin}_{s}\|s_{obs} - s\|^2\hspace{0.1cm}\text{s.\,t.}\, \mathop{det}(J(s)^TJ(s))=0
\label{eq:PIOC}
\end{equation}
\end{small}
\subsection{Problem statement}
Despite the existence of a formula for the gradient of a determinant, the PIOC cannot be solved using gradient based methods such as IPM because it violates the LICQ. Since the matrix $J(s)^TJ(s)$ is a positive-definite matrix, $det(J(s)^TJ(s))$ is always positive. This means that the $SC$ represents the set of minimums of $det(J(s)^TJ(s))$. Thus the gradient of the equality constraint is identically zero along the $SC$. $\forall s \in SC, \quad \frac{\partial \mathop{det}(J(s)^TJ(s))}{\partial s} = 0$. Rewriting \eqref{eq:kkt3} for the PIOC as in \eqref{eq:PIOC_KKT} makes clearer the difficulty in computing the step direction for the IPM as the system becomes singular. This LICQ violation makes gradient based solvers unsuitable.
Additionally, The PIOC has only been formulated to consider unconstrained DOC problems and needs to be extended to handle real world constrained applications.
{\footnotesize
\begin{equation}
    \begin{pmatrix}
        \frac{\partial^2 \mathcal{L}}{\partial s^2} &  \frac{\partial \mathop{det}(J(s)^TJ(s))}{\partial s}^T \\
        \frac{\partial \mathop{det}(J(s)^TJ(s))}{\partial s} & 0 \\
    \end{pmatrix}\begin{pmatrix}
        \Delta s\\ \Delta \lambda
    \end{pmatrix}
    = -E
    \label{eq:PIOC_KKT}
\end{equation}
}
\section{Mathematical Jargon}\label{sec:math_jargon}
\subsection{Riemannian Optimization}\label{sec:rie_opt}
Riemannian optimization\cite{smith_optimization_2014} is a framework where the search space of the optimization variable is restricted to a Riemannian manifold. The formulation of a constrained Riemannian optimization problem can be seen in  \eqref{eq:rie_opt}.
{\footnotesize
\begin{equation}
\begin{split}
    s^{*} = \operatornamewithlimits{argmin}_{s \in \mathcal{M}} f(s)\hspace{0.2cm}\text{s.\,t.}\, g(s)=0,  h(s)\ge0\\
\end{split}
\label{eq:rie_opt}
\end{equation}
}
\subsubsection*{Manifold \cite{lee_introduction_2012} }
A manifold is a differentiable geometry that locally resembles a plane. Mathematically, a smooth embedded submanifold $\mathcal{M}$ in $\mathbb{R}^m$ is a subset of $\mathbb{R}^m$ such that for all $s \in \mathcal{M}$ there exists a neighbourhood $U \subset \mathbf{R}^m$ of $s$ and a differentiable function $\rho : U \longrightarrow \mathbb{R}^{m-dim(\mathcal{M})}$ with full rank differential such that $\mathcal{M}\cap U = \{z\in U |\hspace{0.15cm}\rho(z) = 0\}$. This means that $\mathcal{M}$ is locally defined by $m - dim(\mathcal{M})$ independent equations.
\subsubsection*{Tangent vector \cite{lee_introduction_2012} }
A vector $v$ is a tangent vector of a manifold $\mathcal{M}$ at $s$ if there exists a differentiable curve $c : \mathbb{R} \longrightarrow \mathcal{M}$ such that $c(0) = s$ and $\frac{d c}{dt}(t=0) = v$. The set of tangent vectors of $\mathcal{M}$ at $s$ is denoted $T_s \mathcal{M}$ and is a linear space of dimension $\mathop{dim}(\mathcal{M})$.
\subsubsection*{Riemannian metric}
A Riemannian manifold $(\mathcal{M}, g)$ is a manifold $\mathcal{M}$ equipped with a Riemannian metric $g$. A Riemannian metric $g : s \mapsto g_s$ is a smooth map that assigns to each point $s \in \mathcal{M}$ an inner product on its tangent space denoted $g_s : T_s \mathcal{M} \times T_s \mathcal{M} \longrightarrow \mathbb{R}$. This metric allows for the definition of norms and distances on the manifold.
 \subsubsection*{Riemannian gradient}
The Riemannian gradient of a function $\mathcal{F} : \mathcal{M} \longrightarrow \mathbb{R}$ at $s \in \mathcal{M}$ is the unique tangent vector $\mathop{grad} \mathcal{F}(s) \in T_s \mathcal{M}$ such that $\forall p \in T_s\mathcal{M},\hspace{0.5em} g_s(\mathop{grad} \mathcal{F}(s), p) = \lim_{t \longrightarrow 0} \frac{\mathcal{F}(c(t)) - \mathcal{F}(c(0))}{t}$ with $c(0) = s$ and $c'(0) = p$. The Riemannian gradient describes the direction of greatest increase of the function similarly to the Euclidean gradient.
\subsubsection*{Parameterization}
A parameterization is a smooth full-rank map  $s : \mathbb{R}^{\mathop{dim}(\mathcal{M})} \longrightarrow \mathcal{M}$. Its partial derivatives $\frac{\partial s}{\partial \gamma}(\gamma)$ form a tangent basis and induce the metric $g_{s(\gamma)}(u,v) =  u^T \frac{\partial s}{\partial \gamma}(\gamma)^T \frac{\partial s}{\partial \gamma}(\gamma) v$. It enables computing the Riemannian gradient via the chain rule $\mathop{grad}\mathcal{F}(s(\gamma))= \frac{\partial \mathcal{F}}{\partial \gamma}(\gamma) = \frac{\partial s}{\partial \gamma}(\gamma)\frac{\partial \mathcal{F}}{\partial s} (s(\gamma))$.  In Riemannian optimization, the problem can be described in terms of the parameters as shown in \eqref{eq:rie_opt_coord}.
\begin{equation}
    \gamma^{*} = \operatornamewithlimits{argmin}_{\gamma \in \mathbb{R}^{\mathop{dim}(\mathcal{M})}} f(s(\gamma))\hspace{0.2cm}\text{s.\,t.}\, g(s(\gamma))=0, h(s(\gamma))\ge0\\
\label{eq:rie_opt_coord}
\end{equation}
\subsubsection*{Transversal intersection}
Transversality describes the angle of intersection between manifolds. It is the opposite of notion of parallelism. The transversal intersection of two submanifolds is also a submanifold\cite{lee_introduction_2012}.

\subsection{Determinantal Varieties}
Consider $\Sigma_{m,n}^r = \{M \in \mathbb{M}_{m,n}|\hspace{0.15cm} rank(M) \le r\}$ the set of matrices whose $(r+1)$-minors are all zero. Since the minors of a matrix are polynomials in matrix coefficient, $\Sigma_{m,n}^r$ is defined by a system of polynomial equations and is an algebraic variety by definition. Algebraic varieties are geometries similar to manifolds except that they allow for singular geometry. Examples of singular geometry are self-intersections, cusps and cone singularity. The set $\Sigma_{m,n}^k$ is called a determinantal variety \cite{artin_determinantal_1985}  of rank $r$ and has singular geometry only where the matrix rank drops, meaning it can be considered a manifold in the neighbourhood of matrices of rank $r$.

\section{Riemannian Inverse Optimal Control}\label{sec:rioc}
In search of an alternative method leveraging the gradient based solvers to speed-up the resolution of IOC, we formulate the problem as a Riemannian optimization task of the form   \eqref{eq:rie_opt}. The approach consists of performing an orthogonal projection onto the solution set of the KKT conditions. For clarity, we first consider DOC problems like \eqref{eq:doc_lin} with only equality constraints $g(s) = 0$. Inequality constraints $h(s) \ge 0$ are addressed separately.
\subsection{Equality constraints only}
\subsubsection{Formulation}
Recall that the SC defined as in \eqref{eq:sc_doc}  is the set of trajectories $s$ verifying  the KKT stationarity condition as in \eqref{eq:kkt}. Let $G$ be the set of trajectories verifying the KKT primal feasibility as in \eqref{eq:eq_ver}. The solution space of the KKT conditions is the intersection $\mathcal{M} = SC \cap G$. Under proper conditions that we will explore, $\mathcal{M}$ can be seen as a smooth manifold of dimension $n_c - 1$.
{\footnotesize
\begin{equation}
    SC = \{s \in \mathbb{R}^{m} \hspace{0.5em} | \hspace{0.5em} \mathop{det}(J(s)^TJ(s)) = 0\}
    \label{eq:sc_doc}
\end{equation}
\begin{equation}
    G = \{s \in \mathbb{R}^{m} \hspace{0.5em}|\hspace{0.5em} g(s) = 0\}
    \label{eq:eq_ver}
\end{equation}
}
This means that the parametrization of of $\mathcal{M}$ using the weight vector $\omega$ is redundant as it has $n_c$ coordinate. However, if we choose to normalize the weight vector using the 1-norm such that $\sum_{k=1}^{n_c} \omega_k = 1$, the last coordinate becomes computable from the others. Thus, we choose the parametrization $\gamma$ of $\mathcal{M}$ such that $\forall i < n_c,\hspace{0.5em} \omega_i = \gamma_i$ and $\omega_{n_c} = 1 - \sum_{k=1}^{n_c-1} \gamma_k$.

Since the aim is to perform an orthogonal projection on $\mathcal{M}$, the objective function will be a distance between the observed trajectory and the trajectory resulting from a DOC  $ \| s_{obs} - \mathcal{D}(\gamma) \|^2$.
As it is the case for the bilevel, a positivity constraint on $\omega > 0$ is added. Since $\omega = \begin{pmatrix}
    \gamma, 1 - \sum_{k=1}^{n_c-1}\gamma_k
\end{pmatrix}$ the positivity constraint is instead placed on $\gamma$ as it is the optimization variable. This gives the final definition of the proposed RIOC shown in \eqref{eq:rioc}.
{\footnotesize
\begin{equation}
    \begin{split}
        \mathop{\mathcal{I}_{\text{RIOC}}}(s_{obs})= &\operatornamewithlimits{argmin}_{\gamma \in \mathbb{R}^{n_c - 1}} \| s_{obs} - \mathcal{D}(\gamma) \|^2_2\\
                &\text{s.\,t.}\,\quad \gamma \ge 0,\quad \sum\nolimits_{k=1}^{n_c-1}\gamma_k \le 1\\
    \end{split}
    \label{eq:rioc}
\end{equation}
}
\subsubsection{Conditions for Riemannian Inverse Optimal Control}
Before using the RIOC, several conditions must be met for $\mathcal{M} = SC \cap G$ to be a manifold. We consider conditions that allow for $SC$ and $G$ to be submanifolds and we verify transversality of their intersection during computation.
\begin{itemize}
    \item Condition 1 - Smoothness : Cost functions and constraints must be twice differentiable. If Riemannian Hessians are computed, they must be three times differentiable.
\end{itemize}
For $G$ a submanifold, it is sufficient for $\frac{\partial g}{\partial s}(s) = J_{\lambda}(s)$ to be full rank. This is covered by assuming LICQ on the DOC problem in order for it to be solvable using IPM.
\begin{itemize}
    \item Condition 2 - LICQ : The DOC verifies the LICQ.
\end{itemize}
Providing conditions such that $SC$ is a manifold is more intricate. The strategy is to reason in $\mathbb{M}_{n,m}$ the space of $m \times n$ matrices with $n = n_c + n_g$, t consider once again the transversal intersection of two manifolds. 
The first geometry to consider in  $\mathbb{M}_{n,m}$ is $\mathop{Im}(J) = \{M\in  \mathbb{M}_{n,m}| \exists s \in \mathbb{R}^{m} , M = J(s)\}$ which is the image of the trajectory space by the operator $J(s)$ defined in \eqref{eq:kkt_lin}. In order of $\mathop{Im}(J)$ to be considered a manifold, it is sufficient for the differential of $J$ to be full rank.
\begin{itemize}
        \item Condition 3 - Rank of differential : Let $j(s) = \begin{pmatrix}
            J_0(s)^T & J_1(s)^T & \hdots & J_n(s)^T
        \end{pmatrix}^T$ with $J_i(s)$ be the i-th column of $J(s)$. Then $rank(\frac{\partial j}{\partial s}(s)) = n$.
\end{itemize}
The second geometry to consider is the determinantal variety  $\Sigma_{m,n}^{n-1}$. To consider it a submanifold at the intersection, a condition is added to avoid singular geometry.
\begin{itemize}
    \item Condition 4 - Minimum rank $\hspace{0.15cm} rank(J(s)) \ge n-1$.
\end{itemize}
Finally, every intersecting manifolds must intersect transversaly. The transversality constraints are :
\begin{itemize}
    \item Condition 5 - Transversality 1: $\Sigma_{n-1}$ and $\mathop{Im}(J)$ intersect transversaly.
    \item Condition 6 - Transversality 2: $SC$ and $G$ intersect transversaly.
\end{itemize}
\subsubsection{Computing the Riemannian jacobian}
When conditions are met, the Riemannian jacobian $\frac{\partial \mathcal{D}}{\partial \gamma}$ exists. This section presents a formula for computing the Riemannian jacobian by solving a linear system.

The first step is the KKT conditions shown \eqref{eq:sc_def}. Condition 1 is assumed. Specifically cost functions and constraints must be differentiable. Condition 4 ensures the uniqueness of $\omega$.
{\footnotesize
\begin{equation}
  \begin{cases}
    J_{\omega}(\mathcal{D}(\gamma))\omega(\gamma) + J_{\lambda}(\mathcal{D}(\gamma))\lambda(\gamma) = 0\\
   g(\mathcal{D}(\gamma)) = 0
  \end{cases}
    \label{eq:sc_def}
\end{equation}
}
Assuming condition 2 is verified, $J_{\lambda}$ is non-singular, and $\lambda$ is unique and can be computed via the formula $\lambda = -J_{\lambda}^{+}J_{\omega}\omega$. By replacing $\lambda$ with this expression the first equation becomes $PJ_{\omega}\omega = 0$ with $P=(I-J_{\lambda}J_{\lambda}^{+})$ a projector on the null-space of $J_{\lambda}^T$. Differentiating this equation using the derivative of the pseudo-inverse gives \eqref{eq:diff_stat}.  Cost functions and equality constraints need to be twice differentiable as in condition 1.
{\footnotesize
\begin{equation}
\begin{aligned}
\left(-\frac{\partial J_{\lambda}}{\partial \gamma_k} + J_{\lambda}J_{\lambda}^{+}\frac{\partial J_{\lambda}}{\partial \gamma_k}J_{\lambda}^{+} - J_{\lambda}J_{\lambda}^{+}J_{\lambda}^{+T}\frac{\partial J_{\lambda}}{\partial \gamma_k}^TP\right)J_{\omega}\omega &\\
    + P\frac{\partial J_{\omega}}{\partial \gamma_k} \omega
    + PJ_{\omega} \frac{\partial \omega}{\partial \gamma_k}&= 0\hspace{-10cm}
\end{aligned}
\label{eq:diff_stat}
\end{equation}
}
Replacing $PJ_{\omega}\omega = 0$ into \eqref{eq:diff_stat} simplifies it to \eqref{eq:diff_stat2}. Factorizing and substituting in $\lambda = -J_{\lambda}^{+}J_{\omega}\omega$  gives \eqref{eq:diff_stat4}.
{\footnotesize
\begin{equation}
\begin{split}
   -P\frac{\partial J_{\lambda}}{\partial \gamma_k} J_{\lambda}^{+}J_{\omega}\omega 
    + P\frac{\partial J_{\omega}}{\partial \gamma_k} \omega
    + PJ_{\omega} \frac{\partial \omega}{\partial \gamma_k}&= 0
\end{split}
    \label{eq:diff_stat2}
\end{equation}
\begin{equation}
  P\left ( \frac{\partial J_{\lambda}}{\partial \gamma_k} \lambda + \frac{\partial J_{\omega}}{\partial \gamma_k} \omega \right)= -PJ_{\omega} \frac{\partial \omega}{\partial \gamma_k}
    \label{eq:diff_stat4}
\end{equation}
}

The left-hand side is simplified via chain rule as in \eqref{eq:dev_jwjl}. 
{\footnotesize
\begin{equation}
\begin{split}
    \frac{\partial J_{\lambda}}{\partial \gamma_k} \lambda + \frac{\partial J_{\omega}}{\partial \gamma_k} \omega &=  \sum_{j=1}^{m}\left(  \frac{\partial J_\lambda}{\partial s_j}\lambda + \frac{\partial J_{\omega}}{\partial s_j}\omega\right)\left(\frac{\partial \mathcal{D}}{\partial \gamma_k}\right)_j\\
    &=\underbrace{\left(\sum_{i=1}^{n_c}\omega_i \frac{\partial^2 C_i}{\partial s^2} + \sum_{i=1}^{n_g}\lambda_i \frac{\partial^2 g_i}{\partial s^2} \right)}_M\frac{\partial \mathcal{D}}{\partial \gamma_k}
    \label{eq:dev_jwjl}
\end{split}
\end{equation}
}
Substituting in \eqref{eq:diff_stat4} gives the simplified derivative of the system \eqref{eq:sc_def} shown in \eqref{eq:dsdg_var}. In this case $\frac{\partial \omega}{\partial \gamma} = \begin{pmatrix}
        I_{n_c-1}\\
        -1_{n_c-1}^T\\
    \end{pmatrix}$.
{\footnotesize
\begin{equation}
  \begin{cases}
P M \frac{\partial \mathcal{D}}{\partial \gamma}= -P J_{\omega}\frac{\partial \omega}{\partial \gamma} \\
   J_{\lambda}^T \frac{\partial\mathcal{D}}{\partial \gamma} = 0
  \end{cases}
    \label{eq:dsdg_var}
\end{equation}
}
 The operator $P$ can be decomposed as $P=NN^T$ with $N$ containing a basis of the null-space of $J_{\lambda}^T$ as columns. $N$ is computed from the SVD $J_{\lambda} = U S V^T$ such that $N = U_{n_g:m}$.  Multiplying the first equation by $N^T$ we get $N^T M \frac{\partial \mathcal{D}}{\partial \gamma}= -N^T J_{\omega}B$. When conditions 3 and 5 are met, the dimension of the solution space of this equation must match the dimension of  the tangent space of $SC$. Since $N^T$ and $B$ are full rank, then $rank(-N^TJ_{\omega}B) = rank(J_{\omega}) = n_c - 1$. This implies $\mathop{rank}(N^TM)=m - n_{g}$ in order for the number of independent solutions to be $Dim(SC) = n - 1 = n_c + n_g - 1$ . Thus $\mathop{rank}(N^TM)$ can be computed to verify conditions 5. 

The second equation of \eqref{eq:dsdg_var} indicates that $\frac{\partial \mathcal{D}}{\partial \gamma}$ lies in the null-space of $J_{\lambda}^T$.  Meaning that $\frac{\partial \mathcal{D}}{\partial \gamma} = NX$ with $X$ an $m-n_g \times n_c - 1$ matrix.  This gives \eqref{eq:projection_eq}. 
{\footnotesize
\begin{equation}
    N^TM(NX)= -N^TJ_{\omega}\frac{\partial \omega}{\partial \gamma}
\label{eq:projection_eq}
\end{equation}
}
Assuming condition 6, the tangent space of $\mathcal{M}$ is  $n_c - 1$ dimensional. The dimension of the solution space of \eqref{eq:projection_eq} must reflect it. Since $rank(-N^TJ_{\omega}B) = n_c - 1$ then $N^{T}MN$ must be full rank $\mathop{rank}(N^{T}MN)= m - n_{g}$. This rank can be used to verify condition 6. Solving this system gives the final formula for the Riemannian jacobian in \eqref{eq:rie_grad}. In practice, solving \eqref{eq:projection_eq} is done using linear solvers specialized in symmetric real systems since for better numerical stability.
{\footnotesize
\begin{equation}
\begin{split}
\frac{\partial \mathcal{D}}{\partial \gamma} &= -N(N^{T}MN)^{-1}N^{T}J_{\omega}\frac{\partial \omega}{\partial \gamma}\\
\end{split}
\label{eq:rie_grad}
\end{equation}
}

If the DOC has no constraints, the KKT conditions simplify to only the stationarity condition. In this case the formula for computing the Riemannian jacobian is $\frac{\partial \mathcal{D}}{\partial \gamma} = -M^{-1}J_{\omega}B\\$, and the measure of the transversality for Condition 5 is the invertibility of $M$. Conditions 2 and 6 are not needed.
\subsection{Inequality constraints and variable bounds}
\begin{figure}
    \centering
    \includegraphics[width=0.25\textwidth]{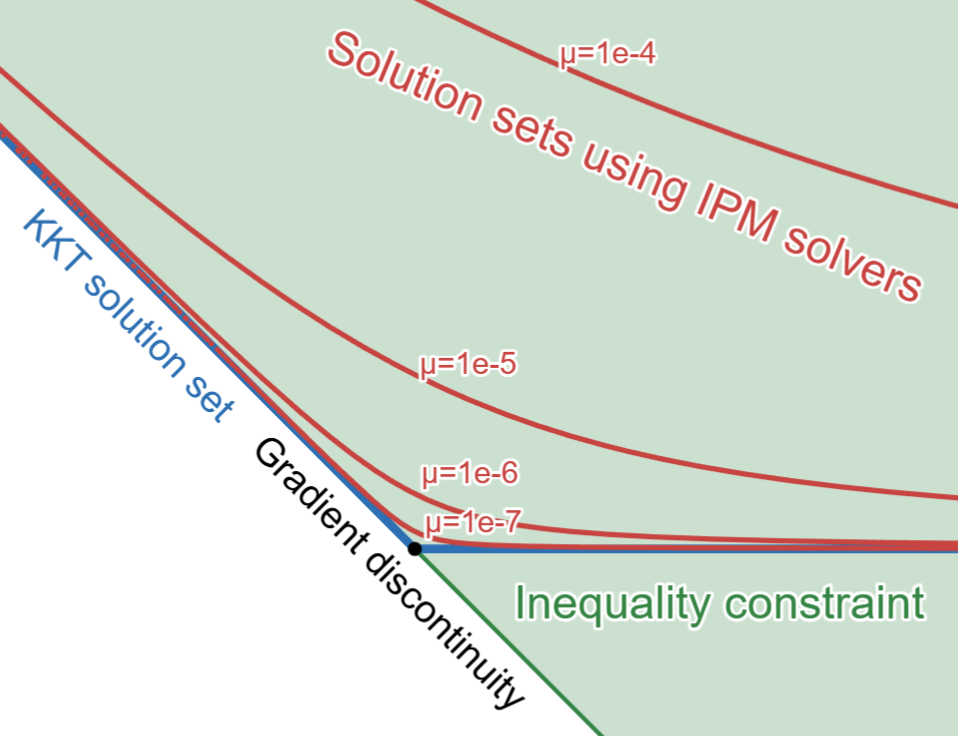}
    \caption{Representation of solution sets of optimization with inequality constraint using IPM solvers for different values of $\mu_{\text{target}}$}
    \label{fig:IPM_set}
\end{figure}
DOC problems with inequality constraints can be viewed as separate equality-constrained DOC problems by treating active inequalities as equalities. However, transitions between active and inactive constraints introduce discontinuities in the Riemannian gradient near the transition point. This effect can be observed in the sudden change in the direction of the KKT solution manifold shown in Fig.~\ref{fig:IPM_set}. 
In practice, IPM solvers avoid this issue by introducing a barrier function $F_{\text{barrier}}(s) = -\sum_{i=1}^{n_h} \mu_{\text{target}} \ln{h_i(s)}$. The KKT multipliers are then computed as $\mu_i(s) = \frac{\mu_{\text{target}}}{h_i(s)}$, where $\mu_{\text{target}}$ is an optimization parameter. This formulation creates a smooth transition between solutions with active and inactive constraints (Fig.~\ref{fig:IPM_set}) and allows the Riemannian Jacobian to be computed without explicitly determining the active set, as in \eqref{eq:sc_def_mu}.
{\footnotesize
\begin{equation}
  \begin{cases}
    J_{\omega}(\mathcal{D}(\gamma))\omega(\gamma) + J_{\lambda}(\mathcal{D}(\gamma))\lambda(\gamma)+  \frac{\partial F_{\text{barrier}}}{\partial s}(\mathcal{D}(\gamma)) = 0\hspace{-0.5cm}\\
   g(\mathcal{D}(\gamma)) = 0
  \end{cases}
    \label{eq:sc_def_mu}
\end{equation}
}
By considering $M_{\mu} = M +\frac{\partial^2 F_{\text{barrier}}}{\partial s^2} $, solving this system is analogous to \eqref{eq:dsdg_var} as in \eqref{eq:rie_grad_mu}. The first equation is multiplied by $\mu_{\text{target}}$ is for numerical stability as computing $\frac{\partial^2 F_{\text{barrier}}}{\partial s^2} = \sum_{i=1}^{n_h}( \frac{\mu_{\text{target}}}{h_i(s)^2}\frac{\partial h_i}{\partial s}(s)^T\frac{\partial h_i}{\partial s}(s) - \frac{\mu_{\text{target}}}{h_i(s)}\frac{\partial^2 h_i}{\partial s^2}(s))$ blows up  for small values of $\mu_{\text{target}}$ because of the term $\frac{\mu_{\text{target}}}{h_i(s)^2} = \frac{\mu_i^2}{\mu_{\text{target}}}$. Computing $\frac{\mu_{\text{target}}^2}{h_i(s)^2} = (\frac{\mu_{\text{target}}}{h_i(s)})^2 =\mu_i^2$ is more stable.
{\footnotesize
\begin{equation}
\begin{split}
\frac{\partial \mathcal{D}}{\partial \gamma} &= -\mu_{\text{target}}N(N^{T}\mu_{\text{target}}M_{\mu}N)^{-1}N^{T}J_{\omega}\frac{\partial \omega}{\partial \gamma}\\
\end{split}
\label{eq:rie_grad_mu}
\end{equation}
}
 
\section{Application}
\label{sec:simulation}
\begin{table}[t]
    \footnotesize
\begin{tabular}{|p{3cm}|c|c|}
 \hline
 Criterion&Symbol&Expression\\
 \hline
 Trajectory& $s$&$(\theta, \dot{\theta}, \tau)$\\
 \hline
  Angular velocity&$C_1$&$\sum_{i=0}^{T} \dot{\theta}_i^T\dot{\theta}_idt$\\
 \hline
 Angular acceleration&$C_2$&$\sum_{t=0}^{T} \ddot{\theta}_i^T\ddot{\theta}_idt$\\
 \hline
 Distance to bar&$C_3$&$\sum_{i=0}^{T} (P_{xi}- P_{x\text{target}})^2dt$\\
 \hline
 End effector velocity&$C_4$&$\sum_{i=0}^{T} \dot{P}_i^T\dot{P}_idt$\\
 \hline
 End effector acceleration&$C_5$&$\sum_{i=0}^{T} \ddot{P}_i^T\ddot{P}_idt$\\
 \hline
 Torque&$C_6$&$\sum_{i=0}^{T} \tau_i^T\tau_idt$\\
 \hline
 Power&$C_7$&$\sum_{i=0}^{T} (\dot{\theta}_i\circ\tau_i)^T(\dot{\theta}_i\circ\tau_i)dt$\\\hline
 Initial position& \multirow{4}{*}{$g$}&$\theta_0 - \theta_{\text{initial}}$\\\cline{1-1}\cline{3-3}
 Initial velocity& &$\dot{\theta}_0$\\\cline{1-1}\cline{3-3}
 Final position& &$P_{x\text{final}} - P_{x\text{target}}$\\\cline{1-1}\cline{3-3}
 Euler dynamics& &
 $F_{\text{Dyn}}(\theta_{i+1}, \dot{\theta}_{i+1}, \theta_{i}, \dot{\theta}_{i}, \tau_{i}, dt)$
\\
 \hline
 Joint limits& \multirow{2}{*}{$h$}&$\theta_i - \theta_{\text{min}}, \quad \theta_{\text{max}} -  \theta_i$\\\cline{1-1}\cline{3-3}
 Torque limits& &$\tau_i - \tau_{\text{min}}, \quad \tau_{\text{max}} - \tau_i$\\\hline
\end{tabular}
    \caption{Direct optimal control formulation}
    \label{tab:doc_for}
\end{table}

\begin{figure}[h]
    \centering
        \includegraphics[width=0.42\textwidth]{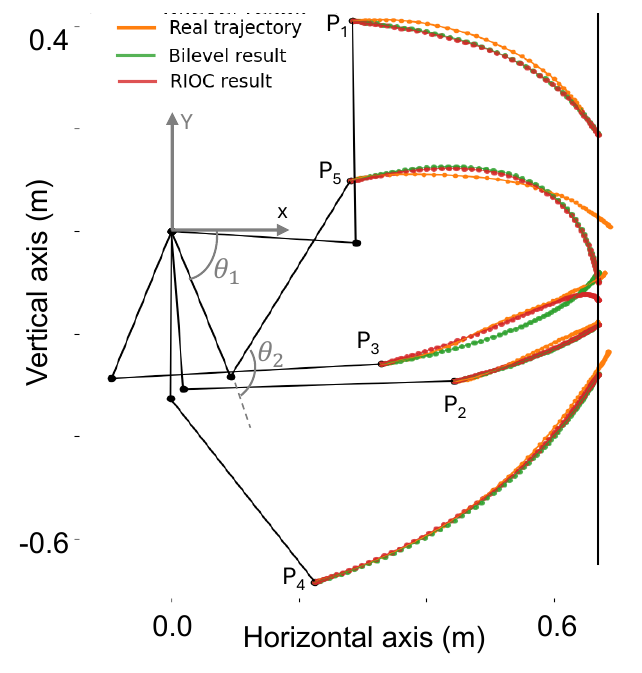}       
    \caption{Cartesian comparaison betwwen human trajectory, bilevel and RIOC for the 5 initial postures. }
    
    \label{fig:results} 
\end{figure}

To compare RIOC with the bilevel approach, both methods were applied to a well-known human arm pointing task \cite{berret_evidence_2011}, for which the authors kindly provided measurement data. A planar two degrees-of-freedom model was used to represent the human arm (Fig.~\ref{fig:results}). Table~\ref{tab:doc_for} describes the DOC (Eq. \ref{eq:doc_lin}) formulated as a constraint multiple shooting problem.
The two IOC methods were used to recover the cost weights from five trajectories of a randomly selected subject. Each trajectory started from a different initial position (P1–P5). Trajectory similarity was measured classically using the Root Mean Square Error on the joint angles. 

\begin{figure}[H]
    \centering

        \includegraphics[width=0.45\textwidth]{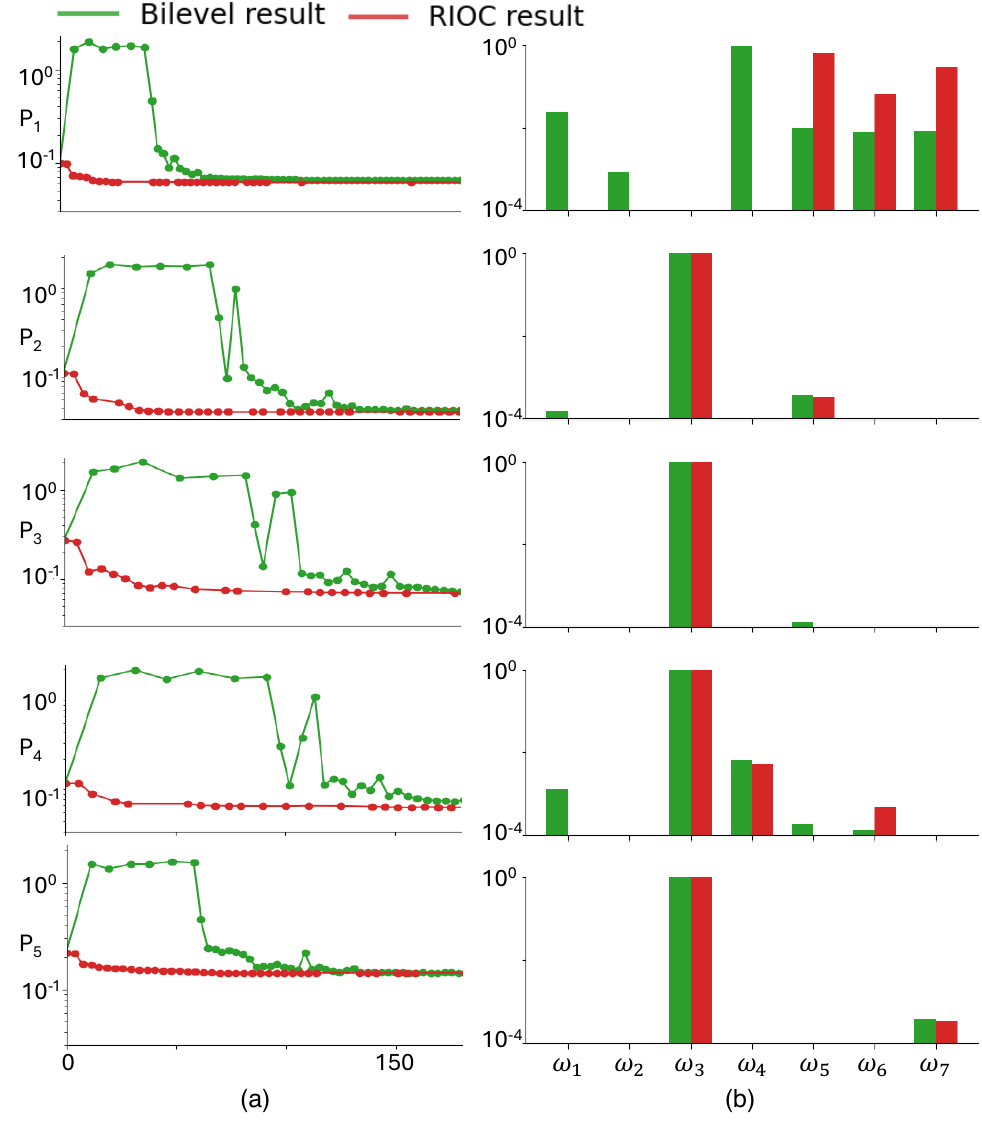}
            
    \caption{(a) Area distance to the observed trajectory versus computation time for the bilevel and RIOC. (b) Corresponding recovered weights.}
    
    \label{fig:results_ioc} 
\end{figure}

Both methods were given a runtime of 3 minutes. The comparison in terms of computation time and accuracy is shown in Fig.~\ref{fig:results_ioc}.a, while the recovered weights are reported in Fig.~\ref{fig:results_ioc}.b. Fig.~\ref{fig:results} illustrates the resulting Cartesian motions obtained with the bilevel method, RIOC, and the measured trajectory.Fig.~\ref{fig:results_ioc}.b shows that the proposed method consistently achieves similar or slightly better precision than the bilevel approach while requiring significantly less computation time. On average, the bilevel method requires $100.2\,\text{s} \pm 50.5$, whereas the proposed method converges in $26.7\,\text{s} \pm 23.9$.
Fig.~\ref{fig:results}.a shows that, in most cases, the recovered trajectories closely match the observed ones in Cartesian space for both the bilevel and RIOC. At the joint level the average RMSE was $4.2 \pm 2.2 \text{deg}$ and  $4.4 \pm 2.2 \text{deg}$ for RIOC and the bilevel, respectively. The main exception was for trajectory P5, where both methods exhibit noticeable discrepancies $8.7 \pm 1.0 \text{deg}$ and $19.9 \pm 26.0 \text{deg}$, likely reflecting limitations in the chosen cost function basis rather than the resolution method. Note that RIOC reduced by 2 the prediction error. Regarding the recovered weights, both methods identify the distance to the bar as the dominant cost function for all trajectories except P1. For P1, the differing and relatively large weight distributions suggest the presence of multiple local minima. In this case, the trajectory reconstructed with the proposed method remains closer to the observed trajectory than the one obtained with the bilevel approach.

\section{Conclusion}\label{sec:conclusion}
This work revisits IOC from a geometric perspective. We show that projection-based IOC formulations suffer from numerical difficulties because the singularity constraints defining optimal trajectories violate standard constraint qualifications. By analyzing the structure of this solution set, we demonstrate that it naturally forms a manifold under appropriate regularity conditions.

Based on this insight, we propose a Riemannian formulation of IOC that performs the trajectory projection directly on this manifold. The resulting RIOC method preserves the advantages of projection-based IOC while enabling reliable gradient-based optimization. Experiments on human arm motion data show that the proposed approach achieves comparable or improved reconstruction accuracy while reducing computation time by nearly 4 compared to classical bilevel IOC.

More broadly, this work highlights the benefit of exploiting the geometric structure of optimality conditions in inverse problems. Future work will investigate extensions to richer biomechanical models and explore how cost function normalization and feature design influence the geometry of the IOC solution space and the interpretability of the recovered costs.
Another promising direction concerns the analysis of cost function identifiability. The derivation of the Riemannian Jacobian relies on the KKT system and on the matrix M describing the sensitivity of the optimal trajectory with respect to the cost parameters. This matrix naturally provides information about how variations of the cost weights affect the resulting trajectory. Studying its conditioning or rank properties could therefore reveal redundancies in the cost function basis. Such sensitivity analysis could be used to automatically identify a minimal and well-conditioned subset of cost functions, improving both numerical stability and interpretability of IOC models.

{\tiny
\printbibliography
}
\end{document}